\documentclass[runningheads]{llncs}

\usepackage{eccv}

\usepackage{eccvabbrv}

\usepackage{graphicx}
\usepackage{booktabs}

\usepackage[pagebackref,breaklinks,colorlinks,citecolor=eccvblue]{hyperref}

\usepackage{orcidlink}

\usepackage{multirow}
\usepackage{pdfpages}
\usepackage{multirow}
\usepackage{wrapfig}

\begin{document}

\title{A Comparative Study of Image Restoration Networks for General Backbone Network Design} 

\titlerunning{A Comparative Study of Image Restoration Networks}

\author{
Xiangyu Chen\textsuperscript{\rm 1,2,3*} Zheyuan Li\textsuperscript{\rm 2,1*} Yuandong Pu\textsuperscript{\rm 3,4*} Yihao Liu\textsuperscript{\rm 2,3} \\Jiantao Zhou\textsuperscript{\rm 1\dag} Yu Qiao\textsuperscript{\rm 2,3} Chao Dong\textsuperscript{\rm 2,3,5\dag}
}

\authorrunning{X.Chen et al.}

\institute{$^{1}$University of Macau $^{2}$Shenzhen Institute of Advanced Technology, \\Chinese Academy of Sciences $^{3}$Shanghai Artificial Intelligence Laboratory \\$^{4}$Shanghai Jiao Tong University $^{5}$Shenzhen University of Advanced Technology\\
\small\url{https://github.com/Andrew0613/X-Restormer}
}

\maketitle

{
\renewcommand{\thefootnote}{}
\footnotetext[1]{* Equal contributions, $^\dag$ Corresponding author.}
}

\begin{abstract}
Despite the significant progress made by deep models in various image restoration tasks, existing image restoration networks still face challenges in terms of task generality. 
An intuitive manifestation is that networks which excel in certain tasks often fail to deliver satisfactory results in others. 
To illustrate this point, we select five representative networks and conduct a comparative study on five classic image restoration tasks.
First, we provide a detailed explanation of the characteristics of different image restoration tasks and backbone networks.
Following this, we present the benchmark results and analyze the reasons behind the performance disparity of different models across various tasks.
Drawing from this comparative study, we propose that a general image restoration backbone network needs to meet the functional requirements of diverse tasks.
Based on this principle, we design a new general image restoration backbone network, X-Restormer. 
Extensive experiments demonstrate that X-Restormer possesses good task generality and achieves state-of-the-art performance across a variety of tasks.
%

\end{abstract}

\section{Introduction}
\label{sec:intro}
\begin{figure}[!t]
\centering
\includegraphics[width=0.85\linewidth]{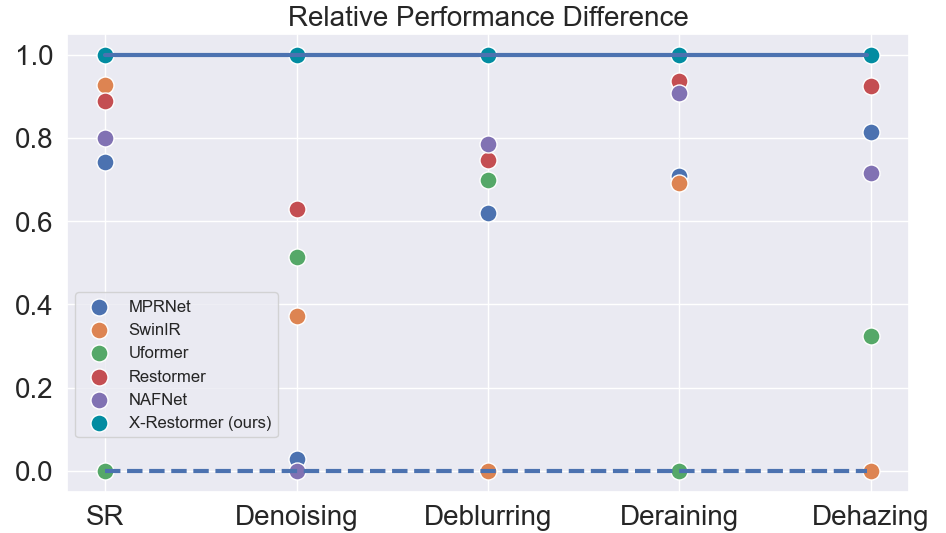}
\caption{Relative performance difference of different backbone networks on five image restoration tasks\protect\footnotemark. The existing representative networks exhibit diverse performance on these tasks, while our method presents superior task generality.}
\label{Distribution}
\end{figure}

Image restoration aims to generate high-quality images from degraded images.
In recent years, deep learning has achieved great success in this field, with numerous networks being proposed to address various image restoration tasks. 
Initially, networks are primarily designed to solve specific restoration tasks and are typically validated only on selected tasks.
As deep learning techniques have continued to evolve, there has been an increasing focus on the development of general-purpose networks that can be applied to a broad range of tasks. 
This trend is particularly evident in the high-level vision field, where new backbone networks are being designed to support multiple tasks~\cite{resnet,swin_t}, including classification, detection and segmentation. 
For image restoration, although more and more backbone networks can handle multiple restoration tasks, their task generality is still limited, as illustrated in \cref{Distribution}. 
For instance, SwinIR~\cite{swinir} achieves state-of-the-art performance on image super-resolution (SR) but falls short on image deblurring and dehazing. 
Conversely, Restormer~\cite{restormer} performs exceptionally well on image dehazing and deraining but is less effective on image SR.
This discrepancy can be attributed to the fact that the characteristics of image degradation vary across different image restoration tasks.
While all image restoration tasks involve mapping degraded images to clean images, the requirements for the capability of backbone networks differ depending on specific tasks. 

\footnotetext{We set the minimum average performance of the networks on test sets in \cref{benchmark} for the task (i) as the lower bound $P^{(i)}_{lower}$, and set the average performance of X-Restormer for each task as the upper bound $P^{(i)}_{upper}$. The ordinate of each point in the figure with performance $P^{(i)}$ is calculated by $(P^{(i)}-P^{(i)}_{lower})/P^{(i)}_{upper}$.}
Designing a general image restoration backbone network presents a significant challenge. However, the development of such a network holds considerable value, as it has the potential to greatly reduce costs associated with research and application.
To achieve this goal, we first conduct a comparative study of mainstream backbone networks on the representative tasks, including image SR, denoising, deblurring, deraining and dehazing. 
These five tasks are chosen due to the distinct characteristics of their degradation. 
%
Five representative backbone networks are selected in the study, including MPRNet~\cite{mprnet}, Uformer~\cite{uformer}, SwinIR~\cite{swinir}, Restormer~\cite{restormer} and NAFNet~\cite{nafnet}. 
These five networks encompass classic architectures such as U-shape architecture, plain residual-in-residual architecture and multi-stage progressive architecture. 
They also employ several common operators, including convolution, spatial self-attention and transposed self-attention~\cite{restormer}. 
We benchmark the five representative methods on the selected five tasks. The experimental results clearly reflect the performance disparity of different backbone networks on different tasks. 
We then conduct a detailed analysis of the characteristics of these tasks and these backbone networks to explain the reasons behind the performance differences. 
Based on the comparative study, we propose that a general backbone network must be highly comprehensive in terms of functionality that meets the diverse needs of various tasks.
%
%

%
It is noteworthy that Restormer stands out in the comparative study, ranking within the top two across all five tasks. 
This superior performance can be attributed to several key designs. 
First, Restormer’s U-shape architecture allows it to process large-size inputs, which is crucial for the tasks that deal with large areas of degradation. Then, the network employs transposed self-attention that utilizes channel-wise features as tokens, achieving the information interaction among channels and enabling the mapping with a global receptive field. Additionally, the incorporation of numerous depth-wise convolutions activates the considerable spatial information interaction ability of the network. From a functional perspective, Restormer integrates the key capabilities of the other compared networks, thereby exhibiting commendable task generality in the comparative study. 
However, the spatial mapping ability of Restormer still appears to be somewhat deficient, as indicated by its quantitatively and qualitatively subpar performance in comparison to SwinIR for SR\footnote{In general, models' SR performance is highly related to the spatial mapping ability.}.
This inferiority is hypothesized to originate from the inherent challenge of detail reconstruction posed by the U-shape architecture, coupled with the relatively weak spatial mapping capability of depth-wise convolution, particularly when compared to spatial self-attention (i.e., window-based self-attention in SwinIR). To address this limitation, a plausible solution is the introduction of spatial self-attention to Restormer.
To achieve this design, we alternately replace half of transposed self-attention blocks with overlapping cross-attention blocks~\cite{hat}, which are proven to have strong spatial information interaction capability, to construct a new network, X-Restormer.
Extensive experiments show that this simple modification can significantly enhance the performance of Restormer without increasing the number of parameters. Moreover, our X-Restormer obtains state-of-the-art performance on all five tasks, exhibiting the best task generality.
Our main contributions can be summarized as follows:
\begin{itemize}
\item[$\bullet$] We conduct a comparative study by constructing an image restoration benchmark, highlighting the challenges faced by existing image restoration backbone networks in task generality. 
\item[$\bullet$] Based on the benchmark results, we perform a detailed analysis of the characteristics of different degradations and networks. We emphasize that the general image restoration backbone network design must meet the functional requirements of diverse tasks. 
\item[$\bullet$] By further enhancing the spatial mapping ability of Restormer, we design a preliminary general backbone network, X-Restormer. Without additional parameters, X-Restormer achieves significant performance improvement over existing networks and exhibits superior task generality. 
\end{itemize}

\section{Related Work}
\label{sec:related_work}
%
\textbf{Image restoration networks.} 
In the past years, numerous deep networks have been proposed for various image restoration tasks such as image SR~\cite{edsr,rdn,hat}, denoising~\cite{dncnn,ircnn,uformer}, deblurring~\cite{dpdnet,hinet}, deraining~\cite{raindata,ipt,tape} and dehazing~\cite{ffanet,maxim,dehazeformer}. 
Initially, most deep networks are designed for specific tasks~\cite{rcan,ffdnet,ddn,dehazenet,Gopro}. 
%
\begin{figure*}[!t]
\centering
\includegraphics[width=0.98\linewidth]{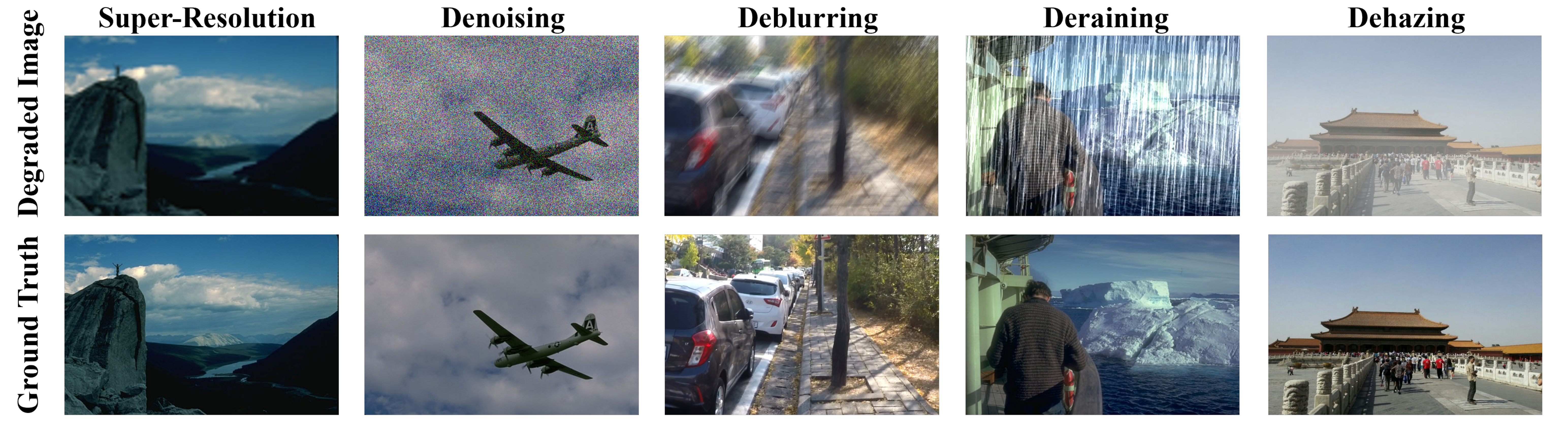}
\caption{Selected five representative image restoration tasks with various degradation.}
\label{IR_tasks}
\end{figure*}
Recently, 
with increasing attention to the task generality of networks, 
more and more methods have been developed to tackle multiple image restoration tasks. For instance, Zamir \etal~\cite{mprnet} builds a multi-stage CNN for deraining, deblurring and denoising. Wang \etal~\cite{uformer} designs a U-shape Transformer for deraining, deblurring and denoising. Liang \etal~\cite{swinir} implements a Swin Transformer-based network that achieves state-of-the-art performance on SR, denoising and compression artifact reduction. Zamir \etal~\cite{restormer} proposes a novel transposed self-attention to build a U-shape network for deraining, deblurring and denoising. Chen \etal~\cite{nafnet} constructs a U-shape CNN for denoising and deblurring.
While existing methods have demonstrated some ability to generalize across several restoration tasks, their task generality remains limited.

%

\noindent\textbf{Difference from the previous network design research.} 
%
%
While previous works have proposed networks that excel in various image restoration tasks, their primary focus is on constructing stronger networks to achieve performance breakthroughs on specific tasks.
In contrast, this work pays more attention to the task generality of the backbone network, possessing a vision different from previous works.
%
%
More specifically, our objective is to explore the design principles and directions of general image restoration networks.
We are not seeking to create powerful networks for peak performance on a single or some specific tasks, but rather to ensure satisfactory performance across a diverse range of tasks.
Regarding the concrete implementation, we do not intend to construct complex network architectures or modules. Our preference, rather, is to enhance task generality through the use of the simplest methodology available.
There are concurrent works that adopt similar ideas for specific image restoration tasks. DAT~\cite{dat} combines spatial-window self-attention and channel-wise self-attention to handle image SR. IPT-V2~\cite{iptv2} designs a spatial-channel Transformer block to build a denoising network and obtains the winner award in the NTIRE 2023 image denoising challenge~\cite{iptv2}. However, the motivation and specific network implementation of our work are distinct from these studies.
%

\section{Image Restoration Benchmark}
\label{sec:benchmark}

In this section, we first briefly introduce several image restoration tasks, each with its own representative degradation characteristics. Subsequently, we classify mainstream image restoration networks based on two key aspects: architecture and core operator. On this basis, we select five representative networks and conduct a benchmark experiment across five different tasks. We describe the experimental setup and explain its rationality. Finally, we present the benchmark results and conduct a detailed analysis of them.

\subsection{Overview of Image Restoration Tasks}

We select five representative tasks for the benchmark experiments. These tasks, exemplified in \cref{IR_tasks}, are chosen based on two primary reasons. 
First, they are very common image restoration tasks with widely accepted evaluation schemes. Second, the degradation characteristics of these tasks are diverse and differ greatly from each other. 
As such, they can provide a robust way to evaluate the task generality of image restoration backbone networks. 
%

%
Let $I_{GT}$ denote the ground truth image and $I_{LQ}$ denote the degraded image, where $I_{GT}\in\mathbb{R}^{H\times W\times 3}$. The degradation model of classic image SR can be represented as:
\begin{equation}
    I_{LQ}=(I_{GT}\otimes k)\downarrow_s,
\end{equation}
where $I_{LQ}\in\mathbb{R}^{\frac{H}{s}\times \frac{W}{s}\times 3}$ represents the low-resolution image. $k$ denotes the bicubic downsampling kernel and $\downarrow_s$ represents the downscaling factor. This degradation is highly correlated to local information and leads to a significant loss of high-frequency information.
Thus, SR networks emphasize strong spatial information interaction capability to reconstruct as many details as possible.
The degradation model of image denoising can be denoted as:
\begin{equation}
    I_{LQ}=I_{GT}+n,
\end{equation}
where $n\in\mathbb{R}^{H\times W\times 3}$ represents the noise map. For Gaussian denoising, noise values are content-independent. 
The downsampling-upsampling process of U-shape architecture inherently aids noise removal. Besides, strong spatial information interaction capability can also enhance high-frequency content reconstruction for denoising networks.
The degradation model of image deblurring (for motion deblurring) can be denoted as:
\begin{equation}
    I_{LQ}=\sum_{t}(f_{motion}^t(I_{GT})),
\end{equation}
where $f_{motion}^t(\cdot)$ represents the motion function under different continuous exposure times. This degradation is related to the global motion offset of the image. Therefore, the ability to utilize large-range information and even global information is important for deblurring networks.
The degradation model of image deraining can be simply denoted as:
\begin{equation}
    I_{LQ}=I_{GT}+R,
\end{equation}
where $R$ denotes the additive rain streak, simulated by the physics models, such as~\cite{li2016rain,liu2018erase}. The difference between this degradation and Gaussian noise is that the added $R$ is not evenly distributed on the image and has a correlation with the image content. Complicated rain streaks also places high demands on the complexity of deraining networks.

\begin{figure*}[!t]
\centering
\includegraphics[width=0.9\linewidth]{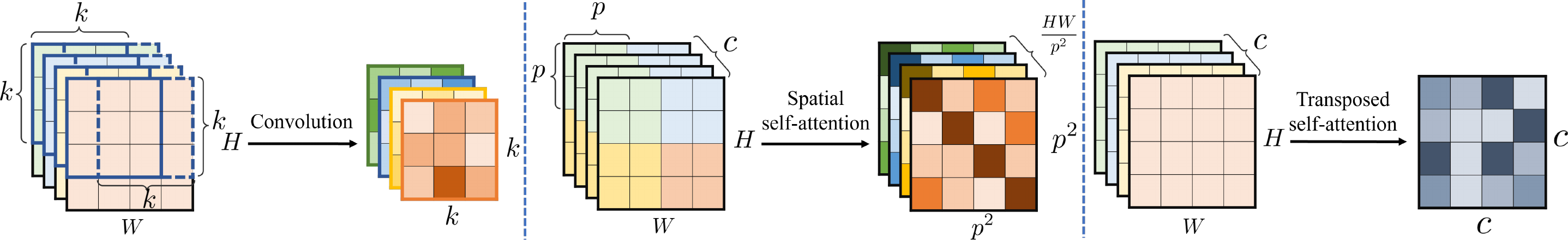}
\caption{The core operators in image restoration networks.}
\label{Operators}
\end{figure*}

The degradation model of image dehazing, based on the atmospheric scattering model, can be denoted as:
\begin{equation}
    I_{LQ}=I_{GT}*t(I_{GT})+A(1-t(I_{GT})),
\end{equation}
where $t(\cdot)$ represents the transmission function and $t(I_{GT})$ is associated with the distance from the scene point to the camera. This degradation is intrinsically linked to the depth information within the image. Consequently, the incorporation of global information is important for dehazing networks.
%

\subsection{Characteristics of Typical Backbone Networks}
The architectures of mainstream image restoration networks can be broadly classified into three categories: U-shape encoder-decoder, plain residual-in-residual and multi-stage progressive. Schematic diagrams of these architectures are provided in \textit{Supp}. 
%
%
The U-shape encoder-decoder architecture performs down-sampling and up-sampling operations on features, enabling networks to handle features of varying scales. This architecture allows networks to accept large-size input, and the effective receptive field of the network expands rapidly with down-sampling. Typical U-shape networks include Uformer~\cite{uformer}, Restormer~\cite{restormer}.
The multi-stage architecture divides the entire network into several sub-networks and progressively processes features, which are primarily used for image deraining and deblurring. Common networks based on this architecture include MPRNet~\cite{mprnet} and HINet~\cite{hinet}.
%
%
The plain residual-in-residual architecture is composed of several residual groups, each of which consists of several residual blocks. This architecture maintains the original size when processing features, which is favorable for the reconstruction of high-frequency information, but it comes at a high computational cost. Typical networks include RCAN~\cite{rcan} and SwinIR~\cite{swinir}.

\begin{wraptable}[9]{r}{0.55\textwidth}
\centering
\caption{Architectures and core operators of the five selected backbone networks.}
\label{IR_networks}
\resizebox{0.55\textwidth}{!}{
\begin{tabular}{c|cc}
\hline
\multirow{2}{*}{Network} & \multirow{2}{*}{Architecture} & \multirow{2}{*}{Core operator} \\
                       &                               &                                      \\ \hline
MPRNet                 & Multi-Stage                   & Convolution                          \\
Uformer               & U-Shape                       & Spatial self-attention                    \\
SwinIR                & Plain residual-in-residual    & Spatial self-attention                    \\
Restormer              & U-Shape                       & Transposed self-attention                 \\
NAFNet                 & U-Shape                       & Convolution                          \\ \hline
\end{tabular}
}
\end{wraptable}

The core operators for constructing an image restoration network can be mainly categorized into three types: convolution, spatial self-attention and transposed self-attention. These operators are shown in \cref{Operators}. The convolution calculates a fixed-size filter and processes the entire feature map through a sliding window, which is the major component of many networks, such as RDN~\cite{rdn_pami}.
%
Spatial self-attention is typically implemented as window self-attention in image restoration tasks. It calculates the attention matrix within a fixed window size, generating content-aware weights that are functionally similar to a large kernel dynamic filter. This operator has strong local fitting ability and shows superior advantages on SR and denoising~\cite{hat_journal}. 
Transposed self-attention treats the entire feature of each channel as a token to calculate the attention matrix on the channel dimension. This operator directly deals with global features, and when combined with depth-wise convolution, it shows remarkable performance in multiple restoration tasks~\cite{restormer}. 
The selected five representative backbone networks for the benchmark experiment encompass the abovementioned architectures and core operators, as presented in \cref{IR_networks}.

\subsection{Experimental Settings}
For image SR, we conduct experiments on upscaling factor $\times 4$. We use the DF2K dataset (the same as SwinIR~\cite{swinir}) to train models. Low-resolution images are generated from the ground truth images using bicubic downsampling in MATLAB. For U-shape networks, we first up-sample the input low-resolution images through bilinear interpolation. The performance is reported on the Y channel. For denoising, we adopt the DFWB dataset for training. Noisy images are generated by adding Gaussian noise with a noise level of 50. For deblurring, we use the motion deblurring dataset GoPro~\cite{Gopro} to train the models. For deraining, we conduct experiments using the synthetic rain dataset Rain13K and calculate the performance on the Y channel, following Restormer~\cite{restormer}. For dehazing, we use the indoor training set (ITS) of the RESIDE dataset~\cite{RESIDE}, the same as~\cite{dehazeformer}.

To maximize the capability of these networks, we use the official codes and training configurations provided by different methods to train the models\footnote{We tried to train all networks with a unified configuration, but find it unreasonable. The performance of networks may vary greatly with different training configurations and optimization strategies, making it difficult to determine a fair unified setting.
}. 
Note that all models are trained without using any pre-training strategy (e.g., $\times2$ pre-training for SR) or special tricks (e.g., EMA in SwinIR and TLC in NAFNet) for fair comparison. 
In addition, we find that different methods may not use exactly the same test sets and the same metrics calculation in their papers to report performance. Therefore, we retest all models based on exactly the same data and calculate metrics using the popular open-source toolbox BasicSR~\cite{basicsr}.

\begin{table*}[!t]
\centering
\caption{Quantitative results on PSNR(dB) of the benchmark experiments. The best and second-best performance results are in \textbf{bold} and \underline{underline}.}
\label{benchmark}
\resizebox{\linewidth}{!}{
\begin{tabular}{c|cc|cc|cc|cc|c}
\hline
\multirow{2}{*}{Method} & \multicolumn{2}{c|}{SR} & \multicolumn{2}{c|}{Denoising} & \multicolumn{2}{c|}{Deblurring} & \multicolumn{2}{c|}{Deraining} & Dehazing 
\\ \cline{2-10} 
                        & Set14     & Urban100    & CBSD68     & Urban100          & GoPro     & HIDE                & Test100     & Rain100H         & SOTS Indoor \\ \hline
MPRNet                  & 28.90     & 26.88       & 28.48      & 29.71             & 32.66     & \underline{30.96}   & 30.29       & 30.43            &\underline{40.34} \\
SwinIR                  & \textbf{29.07} & \textbf{27.47} & \underline{28.56} & 29.88  & 31.66 & 29.41               & 30.05       & 30.45            & 29.14            \\
Uformer                 & 27.14     & 25.60       & 28.55 & \underline{29.98} & \underline{33.05} & 30.89            & 27.93       & 24.06            & 33.58            \\
Restormer               & \underline{29.06} & \underline{27.32} & \textbf{28.60} & \textbf{30.02} & 32.92 & \textbf{31.22} & \textbf{32.03} & \underline{31.48} & \textbf{41.87} \\
NAFNet                 & 29.03      & 27.00       & 28.52      & 29.65             & \textbf{33.08} & \textbf{31.22} & \underline{30.33} & \textbf{32.83} & 38.97        \\ 
\hline
\end{tabular}}
\end{table*}

\subsection{Benchmark Results}
We present the quantitative results of the benchmark experiments in \cref{benchmark}. (Due to space constraints, complete results are provided in \textit{Supp}.) Several important observations can be made from the results:
%
1) Different networks exhibit varying performance on different tasks. For instance, SwinIR performs best on SR but worst on deblurring and dehazing. Uformer excels on denoising and deblurring but performs poorly on deraining and SR. 2) Networks with U-shape and multi-stage architectures present clear advantages on deblurring and dehazing. 3) MPRNet and NAFNet, which are mainly based on convolution operators, exhibit moderate performance across all tasks without outstanding results. 4) SwinIR, which employs plain architecture and spatial self-attention operators, outperforms other networks by a significant margin on SR. 5) The overall performance of Restormer is outstanding. Except for consistently being weaker than SwinIR on SR, it obtains considerable performance on almost all other tasks.

\subsection{Analysis}
%
In this section, we explain the above observations by analyzing the characteristics of different tasks and backbone networks.
The degradation of SR lies in the compression of local information, resulting in a large loss of high-frequency details. 
Therefore, SR networks often require strong spatial information interaction capability, or even generative capability. The U-shape architecture, which incorporates multiple downsampling operations, may undermine the reconstruction of high-frequency information and intuitively escalates the difficulty of detail reconstruction. 
In contrast, the plain architecture that maintains feature sizes benefits SR. Besides, window self-attention has demonstrated a superior local fitting ability than convolution~\cite{hat}. As a result, SwinIR, which is based on a plain structure and employs spatial self-attention operators, exhibits a distinct advantage on SR. 
Denoising entails smoothing the image to eliminate high-frequency noise and integrating low-frequency information to reconstruct a clear image. This task places no explicit unique requirement for the network, while its performance intuitively benefits from effective spatial information interaction. 
It is conjectured that the high performance of Restormer on denoising can be attributed to its ability to better smooth noise through channel-wise processing, akin to operating in the frequency domain. In contrast, SwinIR and Uformer perform well due to their robust spatial information interaction ability of the spatial self-attention. 
%
%

%
Deblurring (specifically for motion blur here) involves addressing global motion shifts in the image. As a result, the ability to handle large-size inputs and the use of global or multi-scale information are necessary for deblurring networks. Thus, the networks based on the U-shape architecture all perform well on this task. Conversely, SwinIR, which employs the plain architecture and focuses more on local information processing, performs much worse than other networks. 
Similar phenomena can be observed for dehazing. Due to the involvement of the depth information in the haze model, the ability to use large-range or even global information is crucial. Besides, dehazing networks are required to handle low-frequency transformations, including alterations in color and contrast, both of which constitute global mappings. Therefore, SwinIR and Uformer, which rely more on local spatial information interaction, perform poorly on this task. On the contrary, Restormer exhibits exceptional performance.
Deraining is relatively unique in that the rain is unevenly distributed in images, with significant differences between different raindrops and streaks. Thus, there is no clear pattern in the performance of different networks on deraining. Nevertheless, networks with higher complexity present better performance. 

Based on the above results and analysis, we can infer that the acceptable performance of a backbone network on a specific task is predicated on meeting the functional requirements of that task. It is notable that Restormer obtains exceptional task generality. This can be attributed to several factors: 1) The U-shape architecture enables the network to accommodate large-size input. 2) The transposed self-attention allows direct interaction of global information. 3) The presence of depth-wise convolution enables the network to process spatial information effectively. In summary, due to Restormer’s comprehensive functionality, it is capable of meeting the diverse requirements of different tasks. 
%

\begin{figure}[!t]
\centering
\includegraphics[width=1\linewidth]{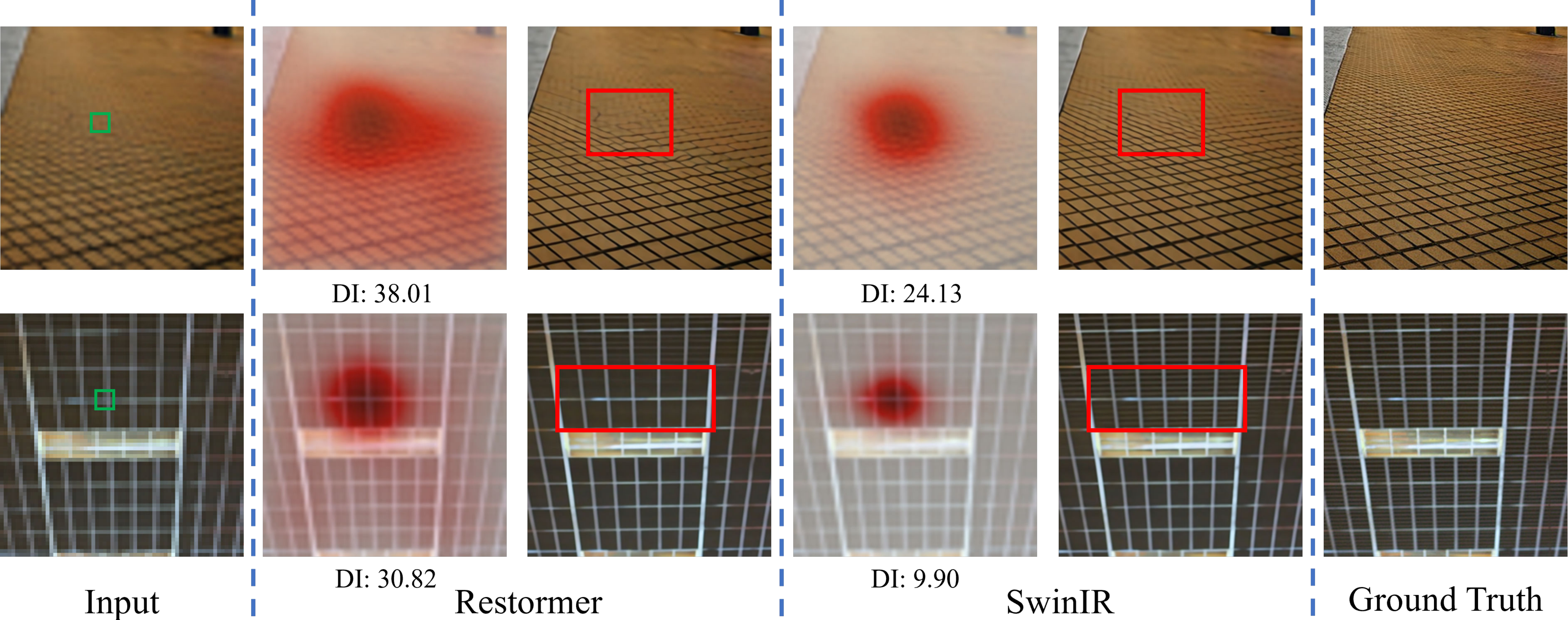}
\caption{Visual and LAM~\cite{lam} comparisons between Restormer and SwinIR. The LAM results and DI values indicate that Restormer exploits significantly more information than SwinIR. However, SwinIR reconstructs much more details than Restormer.}
\label{comparison}
\end{figure}

\begin{figure*}[!t]
\centering
\includegraphics[width=0.98\linewidth]{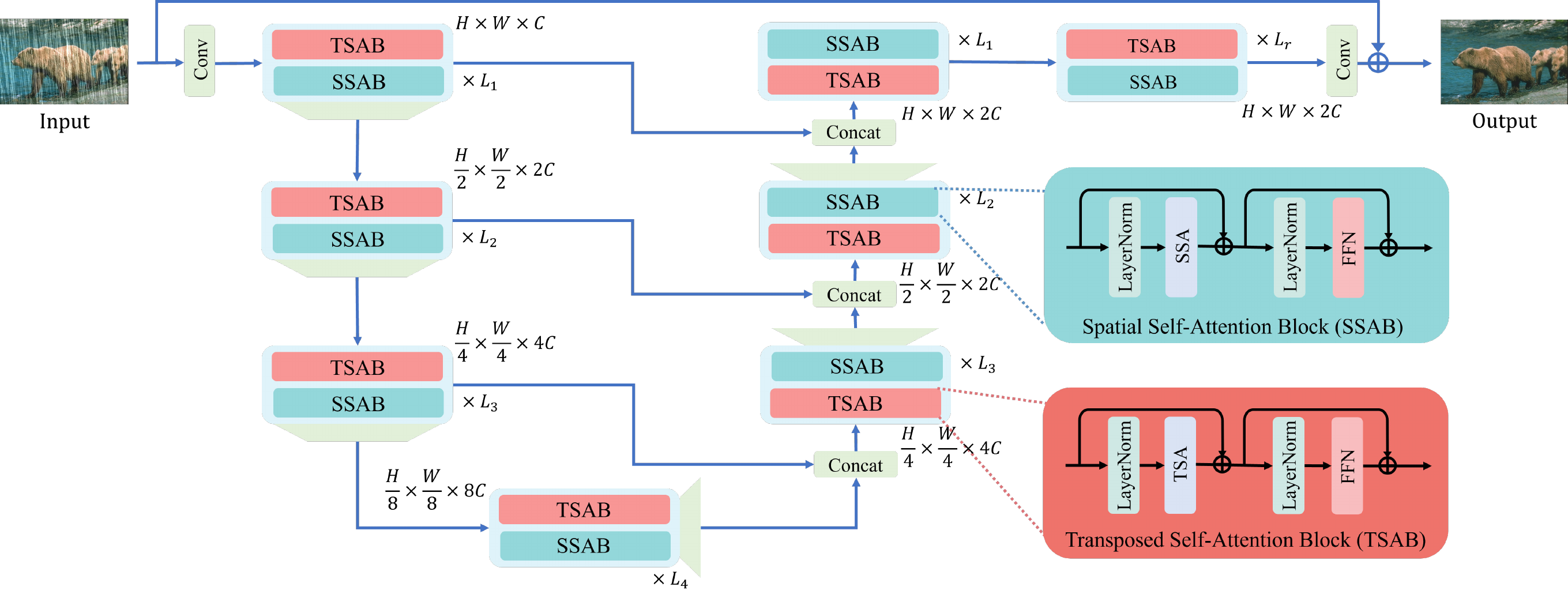}
\caption{The network structure of X-Restormer. To enhance the spatial mapping ability of Restormer and create a more general network, we replace half of the transposed self-attention blocks in Restormer with spatial self-attention blocks. For TSA, we retain the preliminary multi-Dconv head transposed attention (MDTA) used in Restormer. For SSA, we adopt the overlapping cross-attention (OCA) in HAT~\cite{hat}.}
\label{strcuture}
\end{figure*}

\section{General Backbone Network Design}
\label{sec:backbone}

Based on the benchmark experiments, we believe that the principle of designing a general backbone network should be to ensure that the network can fulfill the functional requirements of all tasks. 
As Restormer shows relatively good task generality, we select it as the starting point to design a more general network. 
By pinpointing and addressing the limitation of Restormer, we present an initial version of a general image restoration backbone network in this section.


\textbf{Limitation of Restormer.} In the benchmark experiments, Restormer shows inferior performance to SwinIR on SR, particularly on Urban100. 
The qualitative comparisons also indicate this phenomenon in \cref{comparison}. 
%
%
From the visual and LAM~\cite{lam} results, We can observe that Restormer can exploit large-range and even global information for the reconstruction. 
However, compared to SwinIR, it fails to reconstruct fine textures, even for self-repeated patterns.
%
%
This discrepancy can be attributed to the U-shape architecture adopted by Restormer on the one hand, which increases the difficulty of reconstructing high-frequency information. 
On the other hand, Restormer relies on depth-wise convolution for spatial information interaction, whose spatial mapping capability is relatively weaker than the spatial self-attention in SwinIR.
Considering that the U-shape architecture is indispensable for some tasks, we still need to retain this architectural design for task generality.
To overcome the limitation of Restormer and design a more powerful backbone network, we choose to further enhance its spatial information interaction ability. 
An intuitive and feasible solution is to incorporate the spatial self-attention module into Restormer.


\textbf{Network structure.} In \cref{strcuture}, we present the structure of our proposed backbone network, denoted as X-Restormer. 
We choose the U-shape architecture to build the network.
In contrast to Restormer, we replace half of the transposed self-attention blocks (TSAB) with spatial self-attention blocks (SSAB) to enhance the ability of spatial information interaction. Given an input feature $F_{in}$, the two blocks process it alternately as:
\begin{gather}
F_t = F_{in} + TSA(LN(F_{in})), \\
F_{t\underline{~}out} = F_t + FFN(LN(F_t)), \\
F_s = F_{t\underline{~}out} + SSA(LN(F_{t\underline{~}out})), \\
F_{out} = F_s + FFN(LN(F_s)),
\end{gather}
where $F_t$, $F_{t\underline{~}out}$, $F_s$ and $F_{s\underline{~}out}$ represent the intermediate feature in TSAB, the output of TSAB, the intermediate feature in SSAB and the output of SSAB. $F_{out}$ means the output of the two consecutive blocks, and also serves as the input for the following two blocks. $TSA(\cdot)$ and $SSA(\cdot)$ indicate transposed self-attention (TSA) and spatial self-attention (SSA) modules. $LN(\cdot)$ denotes layer normalization and $FFN(\cdot)$ represents the feed-forward network. 
Specifically, we adopt the Multi-Dconv Transpose Attention (MDTA) as the TSA module. It first generates \textit{query} (\textit{Q}), \textit{key} (\textit{K}) and \textit{value} (\textit{V}) by applying $1\times1$ convolutions followed by $3\times3$ depth-wise convolutions. Then, the channel attention matrix of size $\mathbb{R}^{C\times C}$ is calculated by the dot-product of reshaped \textit{Q} and \textit{K} followed by a Softmax function. The schematic of TSA is shown in \cref{Operators}. Finally, the result is generated by the dot-product of the attention matrix and \textit{V}. 
For SSA, we adopt the Overlapping Cross-Attention (OCA) introduced in the HAT model~\cite{hat}. We choose OCA because the shifted window mechanism in SwinIR is not intuitively suitable for our TSA-SSA consecutive blocks, and HAT demonstrates the effectiveness and superiority of OCA. For the specific calculation, \textit{Q} is produced by partitioning the input into non-overlapping windows, while \textit{K} and \textit{V} are generated by partitioning the input into overlapping windows with a manually set overlapping size. Apart from the different window partition methods, the calculation of OCA is essentially identical to that of standard window self-attention.
For FFN, we employ the Gated-Dconv Feed-forward Network (GDFN) architecture, as used in Restormer. Instead of using two $1\times1$ convolutions to construct an MLP, GDFN first processes input features through two $3\times3$ depth-wise convolutions and $1\times1$ convolutions. Then, the resulting features are combined via element-wise multiplication and pass through another $1\times1$ convolution to produce the final output. 
We have also tried multiple design choices for SSAB and TSAB. Experiments can be found in \textit{Supp}. We emphasize that our design of X-Restormer is not to develop novel architectures or modules to improve the performance on certain tasks, but to enhance the task generality of the network according to the principle of general backbone network design through as simple means as possible.

\section{Experiments}
\label{sec:experiments}

\subsection{Experimental Setup}
We conduct experiments of the proposed X-Restormer on the same datasets used in the benchmark experiment. For the network implementation, the network employs a 4-level encoder-decoder with three times down-sampling and up-sampling. To maintain a similar number of parameters as Restormer, from level-1 to level-4 (\ie, $L_1\sim L_4$ in the figure) the numbers of consecutive blocks (containing a TSAB and a SSAB) are [2, 4, 4, 4] and the number of refinement blocks (\ie, $L_r$) is 4. Attention heads in TSA and SSA are both [1, 2, 4, 8], and channel numbers are [48, 96, 192, 384]. For OCA, the window size and the overlapping ratio are set to 8 and 0.5 as in HAT. The channel expansion factor in GDFN is 2.66. The overall parameters are 26.06M, slightly less than Restormer of 26.13M. We adopt the same training settings as Restormer in the benchmark experiment to optimize the model. We use the AdamW optimizer with $\beta_1=0.9$ and $\beta_2=0.99$, utilizing an initial learning rate of $3e^{-4}$. The learning rate decay follows a cosine scheduler with intervals at 92k and 208k iterations, and the total training iterations are 300K. The input patch size is $256\times256$ and the batch size is 32. For data augmentation, we use horizontal and vertical flips. We utilize the $L1$ loss function to train the model. Notably, we do not adopt any training tricks (e.g., $\times2$ SR pretraining or EMA strategy) or testing tricks (e.g., TLC~\cite{tlc}).

\begin{table}[!t]
\centering
\begin{minipage}[t]{0.54\textwidth}
\centering
\caption{Quantitative results on $\times4$ image SR. 
* means the model pretrained on $\times2$ SR.}
\label{tab:SR_results}
\resizebox{\linewidth}{!}{
\begin{tabular}{c|ccccc}
\toprule[1.5pt]
Model & \multicolumn{1}{c}{Set5}   & \multicolumn{1}{c}{Set14}   & \multicolumn{1}{c}{BSD100}  & \multicolumn{1}{c}{Urban100}  & \multicolumn{1}{c}{Manga109}    \\
\midrule[1.5pt]
RCAN                   & 32.63/0.9002                  & 28.87/0.7889                   &  27.77/0.7436                 &  26.82/0.8087                 &  31.22/0.9173                  \\ 
RCAN-it                   & 32.69/0.9007                  & 28.99/0.7922                   &  27.87/0.7459                 &  27.16/0.8168                 &  31.78/0.9217                  \\ 
SwinIR*                   & 32.92/\underline{0.9044}                  & \underline{29.09}/\underline{0.7950}                   &  27.92/0.7489                 &  27.45/0.8254                 &  \underline{32.03}/\underline{0.9260}                  \\
IPT                   & 32.64/-                  & 29.01/-                   &  27.82/-                 &  27.26/-                 &  -/-                  \\
EDT                   & 32.82/0.9031                  & \underline{29.09}/0.7939                   &  27.91/0.7483                 &  27.46/0.8246                 &  32.05/0.9254                  \\
\hline               
NAFNet                & 32.79/0.9010                  & 29.03/0.7919                   &  27.86/0.7463                 &  27.00/0.8112                 &  31.77/0.9216                  \\ 
SwinIR                & 32.88/0.9041       & 29.07/0.7944      &  \underline{27.93}/\underline{0.7490}    &  \underline{27.47}/\underline{0.8258}      &  31.96/0.9255                 \\
Restormer             & \underline{32.94}/0.9039                & 29.06/0.7934                   &  27.91/0.7482                 &  27.32/0.8199                 &  31.96/0.9244                   \\ \hline
\textbf{X-Restormer}                  & \textbf{33.16}/\textbf{0.9058}                 & \textbf{29.17}/\textbf{0.7963}                   &  \textbf{28.00}/\textbf{0.7512}                 &  \textbf{27.66}/\textbf{0.8291}                 &  \textbf{32.38}/\textbf{0.9279}                 \\ 
\hline
\end{tabular}}
\end{minipage}
\hfill
\begin{minipage}[t]{0.45\linewidth}
\centering
\caption{Quantitative results on image denoising with the noise level $\sigma=50$. 
}
\label{tab:Denoising_results}
\resizebox{\linewidth}{!}{
\begin{tabular}{c|cccc}
\toprule[1.5pt]
Model & \multicolumn{1}{c}{CBSD68}   & \multicolumn{1}{c}{Kodak24}   & \multicolumn{1}{c}{McMaster}  & \multicolumn{1}{c}{Urban100}     \\
\midrule[1.5pt]
FFDNet                 &   27.96/-                    & 28.98/-                   &   29.18/-                 &   28.05/-                 \\                       
RNAN                 &   28.27/-                    & 29.58/-                   &   29.72/-                 &   29.08/-                 \\
RDN                 &   28.31/-                    & 29.66/-                   &   -/-                 &   29.38/-                 \\
IPT                  &   28.39/-                    &   29.64/-                   &    29.98/-                 &   29.71/-                 \\
DRUNet                 &  28.51/-                    &  29.86/-                   &   30.08/-                 &  29.61/-                 \\ 
\hline
SwinIR                 & 28.56/0.8118                  & 29.95/0.8221                   &  30.20/0.8489                 &  29.88/0.8861                 \\
Uformer               & 28.55/\underline{0.8130}                  & 29.97/\underline{0.8244}                   &  30.16/0.8485                 &  29.98/\underline{0.8900}               \\
Restormer           &  \underline{28.60}/\underline{0.8130}                  &  \underline{30.01}/0.8237                   &   \underline{30.30}/\underline{0.8517}                 &  \underline{30.02}/0.8898                 \\ \hline
\textbf{X-Restormer}                            &\textbf{28.63}/\textbf{0.8138}          & \textbf{30.05}/\textbf{0.8245}          &  \textbf{30.33}/\textbf{0.8518}        &  \textbf{30.24}/\textbf{0.8928}    \\ 
\hline
\end{tabular}}
\end{minipage}
\end{table}

\begin{table}[!t]
\centering
\begin{minipage}[t]{0.448\linewidth}
\centering
\caption{Quantitative results on image deblurring (motion blur). 
}
\label{tab:Deblurring_results}
\resizebox{\linewidth}{!}{
\begin{tabular}{c|cccc}
\toprule[1.5pt]
Model & \multicolumn{1}{c}{GoPro}   & \multicolumn{1}{c}{HIDE}   & \multicolumn{1}{c}{RealBlur-R}  & \multicolumn{1}{c}{RealBlur-J}     \\
\midrule[1.5pt]
SPAIR                 & 32.06/0.953                  & 30.29/\underline{0.931}                   &  -/-                         &  28.81/0.875                    \\
MIMO-UNet+           & 32.45/\underline{0.957}                  & 29.99/0.930                   &  35.54/0.947                 &  27.63/0.837                    \\ 
IPT           & 32.52/-                  & -/-                   &  -/-                 &  -/-                    \\ 
MPRNet                 & 32.66/\textbf{0.959}                   & 30.96/\textbf{0.939}                    &  35.99/0.952                  &  28.70/0.873                  \\
\hline
Uformer                & 33.05/0.942                   & 30.89/0.920                    &  \underline{36.19}/0.956      &  \textbf{29.09}/\textbf{0.886}         \\
NAFNet                 & \underline{33.08}/0.942      & \underline{31.22}/0.924        &  35.97/0.952                  &  28.32/0.857                     \\
Restormer              & 32.92/0.940                   & \underline{31.22}/0.923                  &  \underline{36.19}/\underline{0.957}   &  \underline{28.96}/\underline{0.879}      \\
 \hline
\textbf{X-Restormer}             & \textbf{33.44}/{0.946}         & \textbf{31.76}/{0.930}          &  \textbf{36.27}/\textbf{0.958}        &  28.87/0.878                     \\ 
\hline
\end{tabular}}

\end{minipage}
\hfill
\begin{minipage}[t]{0.542\linewidth}
\centering
\caption{Quantitative results on image deraining. 
}
\label{tab:Deraining_results}
\resizebox{\linewidth}{!}{
\begin{tabular}{c|ccccc}
\toprule[1.5pt]
Model & \multicolumn{1}{c}{Test100}   & \multicolumn{1}{c}{Rain100H}   & \multicolumn{1}{c}{Rain100L}  & \multicolumn{1}{c}{Test1200} & \multicolumn{1}{c}{Test2800}    \\
\midrule[1.5pt]
PreNet      &    24.81/0.851                  &   26.77/0.858                   &   32.44/0.950                 &    31.36/ 0.911                &    31.75/0.916                   \\
MSPFN       &   27.50/0.876                  &  28.66/0.860                   &  32.40/0.933                 &   32.39/0.916                &   32.82/0.930                   \\
MPRNet      &    30.27/0.897                  &   30.41/0.890                   &   36.40/0.965                 &    32.91/ 0.916               &    33.64/0.938                   \\
SPAIR       &  30.35/0.909                  & 30.95/0.892                   &  36.93/0.969                 &   \underline{33.04}/\underline{0.922}                &  33.34/0.936                   \\ \hline

SwinIR       & 30.05/0.900                  & 30.45/0.895                   &  37.00/0.969                    &  30.49/0.893                &  33.63/\underline{0.940}                   \\
NAFNet        & 30.33/0.910                  & \textbf{32.83}/\textbf{0.914}       &  36.96/0.971                     &   32.58/\underline{0.922}       &  32.15/0.933                   \\
Restormer  & \underline{32.03}/\underline{0.924}      & 31.48/\underline{0.905}                    &  \underline{39.08}/\textbf{0.979}       &  \textbf{33.22}/\textbf{0.927}          &  \textbf{34.21}/\textbf{0.945}                   \\
 \hline
\textbf{X-Restormer} & \textbf{32.21}/\textbf{0.927}         &  \underline{32.09}/\textbf{0.914}         &  \textbf{39.10}/\underline{0.978} &     32.31/0.919          &    \underline{33.93}/\textbf{0.945}               \\ \hline
\end{tabular}}
\end{minipage}
\end{table}

\begin{table}[!t]
\centering
\caption{Quantitative results on image dehazing. 
}
\label{tab:Dehazing_results}
\resizebox{1\linewidth}{!}{
\begin{tabular}{c|ccccc|ccc|c}
\toprule[1.5pt]
Model & PFDN        & FFA-Net     & AECR-Ne     & MAXIM     & DehazeFormer & MPRNet       & NAFNet       & Restormer    & \textbf{X-Restormer}           \\ 
\midrule[1.5pt]
SOTS Indoor            & 32.68/0.976 & 36.39/0.989 & 37.17/0.990 & 39.72/-   & 40.05/\textbf{0.996}  & 40.34/0.994 & 38.97/0.994 & \underline{41.97}/0.994 & \textbf{42.90}/\underline{0.995} \\ \hline
\end{tabular}}
\end{table}

\subsection{Experimental Results}
We compare our X-Restormer with the top three models in the benchmark experiments (based on the same test configurations) as well as several state-of-the-art approaches for each task (based on the reported performance in their papers) in this section. PSNR(dB)/SSIM is provided in following tables. The best and second-best performance results are in \textbf{bold} and \underline{underline}.

\noindent\textbf{Image SR.}
In \cref{tab:SR_results}, we present the quantitative results of $\times4$ SR on five benchmark datasets: Set5~\cite{set5}, Set14~\cite{set14}, BSD100~\cite{b100}, Urban100~\cite{urban100} and Manga109~\cite{manga109}. The state-of-the-art approaches, including RCAN~\cite{rcan}, RCAN-it~\cite{rcanit}, SwinIR~\cite{han}, IPT~\cite{ipt} and EDT~\cite{edt} are compared in this experiment. 
X-Restormer significantly outperforms Restormer by 0.22dB on Set5, 0.34dB on Urban100 and 0.42dB on Manga109. This demonstrates the effectiveness of our design in enhancing the spatial mapping ability of Restormer. Furthermore, X-Restormer surpasses the SOTA method EDT by 0.2dB on Urban100 and 0.35dB on Manga109, indicating the effectiveness of X-Restormer on SR. 
Despite this, we point out that our method still cannot beat the most powerful SR approaches, \eg, HAT. This is due to the inevitable weakening of SR performance for the U-shape architecture. In terms of SR, the plain residual in residual architecture is still more effective.

\noindent\textbf{Image denoising.}
In \cref{tab:Denoising_results}, we provide the quantitative results of Gaussian denoising with the noise level $\sigma=50$ on four benchmark datasets: CBSD68~\cite{bsd68}, Kodak24~\cite{kodak}, McMaster~\cite{mcmaster} and Urban100~\cite{urban100}. The state-of-the-art methods: FFDNet~\cite{ffdnet}, RNAN~\cite{rnan}, RDN~\cite{rdn_pami}, IPT~\cite{ipt} and DRUNet~\cite{drunet} are compared in this experiment. X-Restormer achieves the state-of-the-art performance, surpassing SwinIR by 0.36dB and outperforming Restormer by 0.22dB on Urban100. This demonstrates the superiority of X-Restormer on image denoising. 

\noindent\textbf{Image deblurring.}
In \cref{tab:Deblurring_results}, we compare the results of X-Restormer with the state-of-the-art methods: SPAIR~\cite{spair}, MIMO-UNet+~\cite{mimounet}, IPT~\cite{ipt} and MPRNet~\cite{mprnet} on both synthetic datasets (Gopro~\cite{Gopro} and HIDE~\cite{hide}) and real-world datasets (RealBlur-R and RealBlur-J~\cite{realblur}). X-Restormer achieves large performance gains over the other models on synthetic datasets, with an improvement of 0.36dB on Gopro compared to NAFNet\footnote{By using TLC, on Gopro/HIDE, NAFNet: 33.69/31.32, X-Restormer: 33.89/31.87.} and 0.54dB on HIDE compared to Restormer. Besides, our X-Restormer obtains the state-of-the-art performance on RealBlur-R and considerable performance on RealBlur-J, showing the effectiveness of our method on real-world motion deblurring scenarios. 
%

\noindent\textbf{Image deraining.}
In \cref{tab:Deraining_results}, we present the quantitative results of deraining on Test100~\cite{test100}, Rain100L~\cite{rain100}, Rain100H~\cite{rain100}, Test1200~\cite{test1200} and Test2800\cite{test2800}. The state-of-the-art methods: PreNet\cite{prenet}, MSPFN~\cite{mspfn}, MPRNet~\cite{mprnet} and SPAIR~\cite{spair} are compared. X-Restormer outperforms the other models on Test100, Rain100H and Rain100L but performs inferior to Restormer on Test1200 and Test2800. This discrepancy is due to the variations in degradation produced by different rain models. Nonetheless, X-Restormer exhibits comparable performance to state-of-the-art methods, showing its effectiveness on image deraining.
%

\noindent\textbf{Image dehazing.}
In \cref{tab:Dehazing_results}, we provide the quantitative results on SOTS Indoor~\cite{RESIDE}. We compare the state-of-the-art approaches: PFDN~\cite{pfdn}, FFA-Net~\cite{ffanet}, AECR-Net~\cite{aecrnet}, MAXIM~\cite{maxim} and DehazeFormer~\cite{dehazeformer} in this experiment. Notably, X-Restormer model significantly outperforms Restormer by a large margin of 0.93dB. When compared to the state-of-the-art dehazing method DehazeFormer, our method achieves a breakthrough performance gain of 2.85 dB. These results demonstrate the superiority of X-Restormer for image dehazing. 

\begin{table*}[!t]
\centering
\caption{Quantitative results on All-in-One restoration. 
}
\label{tab:AllinOne_results}
\resizebox{1\linewidth}{!}{
\begin{tabular}{c|cc|ccc|c|c|c}
\toprule[1.5pt]
\multirow{2}{*}{Model} & \multicolumn{2}{c|}{SR} & \multicolumn{3}{c|}{Denoising} & \multirow{2}{*}{Deblurring} & \multirow{2}{*}{Deraining} & \multirow{2}{*}{Dehazing} \\ 
\cline{2-6} 
& $\times2$ & $\times4$ & $\sigma=15$ & $\sigma=25$ & $\sigma=50$ &  &  & \\ 
\midrule[1.5pt]
MPRNet & 33.68/0.9300 & 28.17/0.8043 & 34.27/0.9280 & 31.82/0.8901 & 28.60/0.8119 & 30.00/0.8812 & 31.20/0.9068 & 35.06/0.9874 \\ 
SwinIR & 33.83/0.9301 & 28.14/0.8043 & 34.27/0.9283 & 31.83/0.8906  & 28.59/0.8143 & 29.06/0.8519 & 30.03/0.8983 & 31.48/0.9823 \\ 
Uformer & 29.99/0.8805 & 27.88/0.7949 & 33.86/0.9254 & 31.42/0.8863 & 27.87/0.7891 & 29.64/0.8725 & 27.53/0.8569 & 29.92/0.9714 \\ 
Restormer & \underline{34.51}/\underline{0.9341} & \underline{28.70}/\underline{0.8179} & \underline{34.43}/\underline{0.9303} & \underline{32.02}/\underline{0.8942} & \underline{28.87}/\underline{0.8222} & \underline{30.54}/0.8902 & \underline{31.91}/0.9134 & \underline{36.95}/\underline{0.9897} \\ 
NAFNet & 34.12/0.9314 & 28.17/0.8087 & 34.18/0.9281 & 31.76/0.8908 & 28.64/0.8187 & 30.38/\underline{0.8911} & 31.56/\underline{0.9149} & 30.84/0.9797\\ 
\hline
\textbf{X-Restormer} & \textbf{34.72}/\textbf{0.9360} & \textbf{28.81}/\textbf{0.8217} & \textbf{34.67}/{0.9330} & \textbf{32.26}/{0.8983} & \textbf{29.12}/\textbf{0.8293} & \textbf{30.85}/\textbf{0.8983} & \textbf{32.27}/\textbf{0.9229} & \textbf{38.24}/\textbf{0.9914} \\ 
\hline
\end{tabular}}
\end{table*}

\noindent\textbf{All-in-One Restoration.} 
We conduct experiments on an all-in-one restoration setting to show the effectiveness of different backbone networks in addressing various tasks simultaneously. 
Networks are trained on five tasks with varying degradation levels (i.e., $\times2$, $\times4$ for SR and $\sigma\in(0,50)$ random level for denoising). 
The sampling probability for each task is the same, and the average performance on benchmark datasets is calculated. 
%
%
As shown in \cref{tab:AllinOne_results}, with the relatively better task generality among the existing networks, Restormer exhibits better performance on the all-in-one restoration. 
By overcoming the limitation of Restormer, our X-Restormer demonstrates further advantages in handling multiple tasks concurrently, with its performance far exceeding other networks on all tasks. 
In contrast, the other networks are more or less affected by optimization conflicts across different tasks (\eg, SwinIR performs inferior to Restormer even on SR). 
These indicate that a general backbone network is of great significance for building a general model that process multiple image restoration tasks, which can effectively mitigate task conflicts with the performance drops. 

\noindent\textbf{Summary.} With enhanced spatial mapping capability, our X-Restormer can significantly outperform Restormer. Specifically, X-Restormer obtains performance gains against Restormer of 0.42dB (Manga109), 0.22dB (Urban100), 0.54dB (HIDE), 0.61dB (Rain100H) and 0.93dB (SOTS Indoor) on image SR, denoising, deblurring, deraining and dehazing, respectively, showing the effectiveness of our design. Despite its simplicity, X-Restormer obtains state-of-the-art performance on all these five tasks and present the best task generality among the compared methods. Furthermore, we show that a more general backbone network can also better handle multiple restoration tasks simultaneously. We hope it can inspire more works on the general image restoration backbone network design. 

\section{Conclusion}
In this paper, we conduct a comparative study of existing image restoration backbone networks to design a general backbone network. Five representative networks are chosen for the benchmark experiment across selected five tasks. The results indicate that comprehensive functionality is crucial for designing a general restoration backbone network. We select Restormer as the baseline and introduce spatial self-attention into it to enhance the spatial information interaction capability. Experimental results show that our X-Restormer achieves significant performance improvement and presents the best task generality. 

\section*{Acknowledgements}
This work was partially supported by National Natural Science Foundation of China (Grant No.62276251, 62272450), and the Joint Lab of CAS-HK. This work was also supported in part by Macau Science and Technology Development Fund under SKLIOTSC-2021-2023 and 0022/2022/A. 
\bibliographystyle{splncs04}
\bibliography{main}

\clearpage



\section*{Supplementary material}

# A Comparative Study of Image Restoration Networks for General Backbone Network Design

## Supplementary Material

Xiangyu Chen[*] Zheyuan Li[*] Yuandong Pu[*] Yihao Liu
Jiantao Zhou[†] Yu Qiao Chao Dong[†]

**Abstract.** In this supplementary material, we present additional experiments and results to complement the main manuscript. First, we illustrate the importance of a general backbone network for multi-task optimization on all-in-one restoration. Then, we complement details about schematic diagrams of mentioned three kinds of architectures. Next, we provide complete benchmark results for the five image restoration tasks and present visual results of X-Restormer with other benchmark networks. After that, we conduct comprehensive ablation study to further verify the effectiveness of our X-Restormer. Finally, we present a comprehensive comparison of the model complexity of different networks.

## 1  Alleviating optimization conflicts

In the manuscript, we show that our X-Restormer obtains the best average performance on an all-in-one restoration setting. We provide more explanations in this section to illustrate a more general backbone network can effectively alleviate the optimization conflict for multi-task restoration. In Fig. 1, we provide the validation curves of different models on image SR×4 and deblurring. X-Restormer can consistently outperform other networks during training. Restormer with good task generality also performs the second best. However, SwinIR performs poorly on both SR and deblurring. We believe this is because when a network has difficulty optimizing certain tasks, the larger gradients it generates may cause the network to focus on these more challenging tasks. Conversely, a general backbone network can effectively circumvent such optimization conflicts.

## 2  Architecture Schematic Diagrams

In the realm of image restoration networks, prevailing architectures fall into three distinct categories: (a) plain residual-in-residual, (b) U-shaped encoder-decoder and (c) multi-stage progressive architectures. In the section 3.2 of the manuscript, we detail the characteristics of these three kinds of architectures. In Fig. 2, we provide the schematic diagrams of these architectures for better understanding of their characteristics.

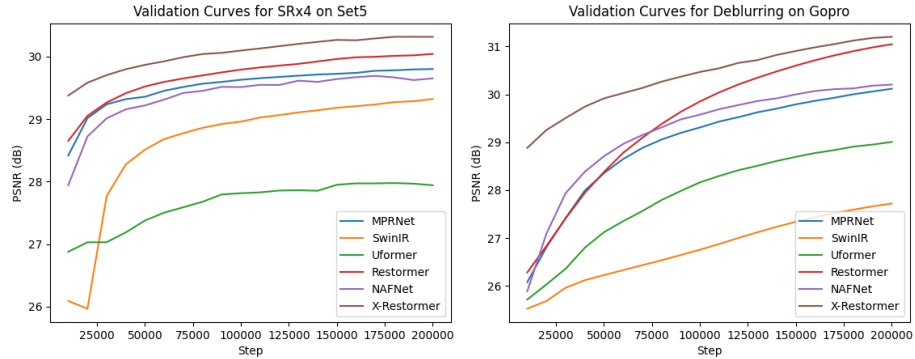

**Fig. 1:** Validation curves for SR×4 and deblurring.

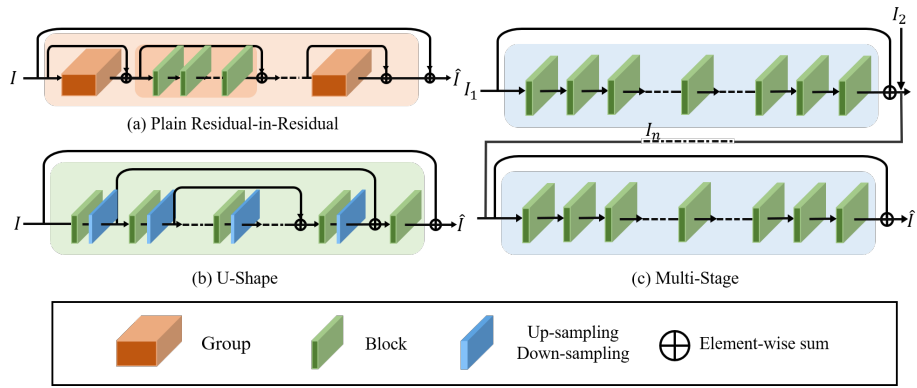

**Fig. 2:** Plain residual-in-residual, U-shape and multi-stage architectures.

## 3   Complete Benchmark Results

We provide the complete the benchmark results on image SR, denoising, deblurring, deraining and dehazing, as shown in Tab. 1, Tab. 2, Tab. 3, Tab. 4 and Tab. 5. Since different methods may not use exactly the same data and calculation ways in their papers, we uniformly calculate the performance using BasicSR toolbox [3] on the totally the same test data for fair comparison.

## 4   Visual Comparison

We present the visual results of the benchmark experiments with X-Restormer on image SR, denoising, deblurring, deraining and dehazing, as depicted in Fig. 3, Fig. 4, Fig. 5, Fig. 6, Fig. 7, respectively. Our X-Restormer obtains the best visual quality compared to other networks.

**Table 1:** Complete benchmark results on ×4 image SR. The best and second-best performance results are in **bold** and <u>underline</u>.

| Model | Set5 PSNR/SSIM | Set14 PSNR/SSIM | BSD100 PSNR/SSIM | Urban100 PSNR/SSIM | Manga109 PSNR/SSIM |
|---|---|---|---|---|---|
| MPRNet | 32.57/0.8996 | 28.90/0.7889 | 27.78/0.7425 | 26.88/0.8081 | 31.44/0.9182 |
| SwinIR | 32.88/0.9041 | <u>29.07</u>/<u>0.7944</u> | <u>27.93</u>/<u>0.7490</u> | <u>27.47</u>/0.8258 | <u>31.96</u>/<u>0.9255</u> |
| Uformer | 30.25/0.8665 | 27.14/0.7398 | 27.67/0.7475 | 25.60/0.7651 | 31.69/0.9233 |
| NAFNet | 32.79/0.9010 | 29.03/0.7919 | 27.86/0.7463 | 27.00/0.8112 | 31.77/0.9216 |
| Restormer | <u>32.94</u>/<u>0.9039</u> | 29.06/0.7934 | 27.91/0.7482 | 27.32/0.8199 | **31.96**/0.9244 |
| **X-Restormer** | **33.16**/**0.9058** | **29.17**/**0.7963** | **28.00**/**0.7512** | **27.66**/**0.8291** | **32.38**/**0.9279** |

**Table 2:** Complete benchmark results on image denoising with the Gaussian noise level $\sigma = 50$. The best and second-best performance results are in **bold** and <u>underline</u>.

| Model | CBSD68 PSNR/SSIM | Kodak24 PSNR/SSIM | McMaster PSNR/SSIM | Urban100 PSNR/SSIM |
|---|---|---|---|---|
| MPRNet | 28.48/0.8087 | 29.86/0.8193 | 30.04/0.8447 | 29.71/0.8847 |
| SwinIR | 28.56/0.8118 | 29.95/0.8221 | 30.20/0.8489 | 29.88/0.8861 |
| Uformer | 28.55/<u>0.8130</u> | 29.97/<u>0.8244</u> | 30.16/0.8485 | 29.98/<u>0.8900</u> |
| NAFNet | 28.52/0.8098 | 29.90/0.8204 | 30.07/0.8455 | 29.65/0.8840 |
| Restormer | <u>28.60</u>/<u>0.8130</u> | <u>30.01</u>/0.8237 | <u>30.30</u>/<u>0.8517</u> | <u>30.02</u>/0.8898 |
| **X-Restormer** | **28.63**/**0.8138** | **30.05**/**0.8245** | **30.33**/**0.8518** | **30.24**/**0.8928** |

**Table 3:** Complete benchmark results on motion deblurring. The best and second-best performance results are in **bold** and <u>underline</u>.

| Model | GoPro PSNR/SSIM | HIDE PSNR/SSIM | RealBlur-R PSNR/SSIM | RealBlur-J PSNR/SSIM |
|---|---|---|---|---|
| MPRNet | 32.66/0.939 | 30.96/0.919 | 35.99/0.952 | 28.70/0.873 |
| SwinIR | 31.66/0.921 | 29.41/0.896 | 35.49/0.947 | 27.55/0.840 |
| Uformer | 33.05/<u>0.942</u> | 30.89/0.920 | <u>36.19</u>/0.956 | **29.09**/**0.886** |
| NAFNet[1] | <u>33.08</u>/<u>0.942</u> | <u>31.22</u>/**0.924** | 35.97/0.952 | 28.32/0.857 |
| SwinIR | 31.66/0.921 | 29.41/0.896 | 35.49/0.947 | 27.55/0.840 |
| Restormer | 32.92/0.940 | <u>31.22</u>/0.923 | <u>36.19</u>/<u>0.957</u> | <u>28.96</u>/<u>0.879</u> |
| **X-Restormer** | **33.44**/**0.946** | **31.76**/**0.930** | **36.27**/**0.958** | 28.87/0.878 |

**Table 4:** Complete benchmark results on deraining. The best and second-best performance results are in **bold** and <u>underline</u>.

| Model | Test100 PSNR/SSIM | Rain100H PSNR/SSIM | Rain100L PSNR/SSIM | Test1200 PSNR/SSIM | Test2800 PSNR/SSIM |
|---|---|---|---|---|---|
| MPRNet | 30.29/0.898 | 30.43/0.891 | 36.46/0.966 | <u>32.94</u>/0.914 | 33.66/0.939 |
| SwinIR | 30.05/0.900 | 30.45/0.895 | 37.00/0.969 | 30.49/0.893 | 33.63/<u>0.940</u> |
| Uformer | 27.93/0.891 | 24.06/0.845 | 35.96/0.965 | 32.75/0.919 | 28.22/0.913 |
| NAFNet | 30.33/0.910 | **32.83**/**0.914** | 36.96/0.971 | 32.58/<u>0.922</u> | 32.15/0.933 |
| Restormer | <u>32.03</u>/**0.924** | 31.48/<u>0.905</u> | <u>39.08</u>/**0.979** | **33.22**/**0.927** | **34.21**/**0.945** |
| **X-Restormer** | **32.21**/**0.927** | <u>32.09</u>/**0.914** | **39.10**/<u>0.978</u> | 32.31/0.919 | <u>33.93</u>/**0.945** |

---

[1] By using TLC, on Gopro/HIDE, NAFNet: 33.69/31.32, X-Restormer: **33.89**/**31.87**.

**Table 5:** Complete benchmark results on dehazing. The best and second-best performance results are in **bold** and <u>underline</u>.

| Model | MPRNet PSNR/SSIM | SwinIR PSNR/SSIM | Uformer PSNR/SSIM | NAFNet PSNR/SSIM | Restormer PSNR/SSIM | X-Restormer PSNR/SSIM |
|---|---|---|---|---|---|---|
| SOTS Indoor | 40.34/0.994 | 29.14/0.968 | 33.58/0.986 | 38.97/0.993 | <u>41.97</u>/<u>0.994</u> | **42.90**/**0.995** |

# 5  Ablation Study

We conduct ablation study to explore the effectiveness of our design for X-Restormer. For fast validation, experiments are implemented on image SR and deblurring tasks with significantly different characteristics, based on a small variant of X-Restormer with the numbers of consecutive blocks of [1,2,2,4], the channel numbers of [36, 72, 144, 288] and the input patch size of $192 \times 192$.

**Cascade or Parallel.** In Tab. 6, we show the quantitative results on image SR and deblurring for the cascade and parallel connections of TSAB and SSAB. The model using cascade design obtains much better performance on test sets. Thus, we use the cascade connection for TSAB and SSAB in X-Restormer.

**Sequences of TSAB and SSAB.** In Tab. 7, we present the quantitative results on image SR and deblurring for different sequences of TSAB and SSAB. SSAB-SSAB means that all TSAB in Restormer are replaced by SSAB. TSAB-SSAB indicates that the features first go through TSAB and then SSAB, while SSAB-TSAB represents the opposite process. SSAB-SSAB performs comparably to the models with half TSAB on SR, while exhibit inferior performance on deblurring. It is reasonable that the global information interaction capability provided by TSAB is important for some tasks. SSAB-TSAB and TSAB-SSAB obtain similar performance on SR, while TSAB-SSAB achieves better performance on image deblurring. Therefore, we use TSAB-SSAB as the default choice.

**Effectiveness of OCA.** In Tab. 8, we provide the quantitative comparison of standard non-overlapping window self-attention (WSA) with the used overlapping self-attention (OCA) for the choice of spatial self-attention in X-Restormer. The overlapping size is set to 0.5, the same as HAT [1]. We can see that OCA performs better than WSA. Therefore, we adopt OCA in X-Restormer.

**Different choices for TSAB and SSAB.** In Tab. 9, we conduct ablation study on different choices for TSAB and SSAB. MDTA [4] and CAB [1,6] are two modules that involve channel-wise mapping. Swin [2], Dilation self-attention [5] and OCA [1] are three prevalent options for spatial self-attention. We implement this experiment based on the comparable number of parameters for all models. When employing MDTA as the choice for TSAB, the model using OCA performs the best compared to other options for SSAB. When adopting OCA as the choice for SSAB, the model with CAB obtains comparable performance to MDTA+OCA on SR, while it performs inferior on image deblurring. We believe this is because CAB contains many convolutions, which can enhance the spatial mapping ability of the model. However, its capability of global information interaction is weaker than MDTA using channel self-attention. Thus, we use the combination of MDTA and OCA as our default choice.

**Table 6:** Ablation study on the connection ways.

| Model | Set5 PSNR/SSIM | Set14 PSNR/SSIM | Urban100 PSNR/SSIM | GoPro PSNR/SSIM | HIDE PSNR/SSIM |
|---|---|---|---|---|---|
| Parallel | 32.61/0.9002 | 28.92/0.7902 | 26.85/0.8083 | 31.53/0.9222 | 29.67/0.9016 |
| Cascade(ours) | **32.82/0.9023** | **29.05/0.7925** | **27.17/0.8158** | **32.25/0.9316** | **30.50/0.9136** |

**Table 7:** Ablation study on different sequences of TSAB and SSAB.

| Model | Set5 PSNR/SSIM | Set14 PSNR/SSIM | Urban100 PSNR/SSIM | GoPro PSNR/SSIM | HIDE PSNR/SSIM |
|---|---|---|---|---|---|
| SSAB-SSAB | 32.76/0.9017 | **29.05/0.7931** | **27.20/0.8178** | 32.22/0.9311 | 30.20/0.9098 |
| SSAB-TSAB | **32.82**/0.9021 | **29.05**/0.7927 | 27.19/0.8162 | 32.13/0.9299 | 30.45/**0.9137** |
| TSAB-SSAB | **32.82/0.9023** | **29.05**/0.7925 | 27.17/0.8158 | **32.25/0.9316** | **30.50**/0.9136 |

**Table 8:** Ablation study on the effectiveness of OCA.

| Model | Set5 PSNR/SSIM | Set14 PSNR/SSIM | Urban100 PSNR/SSIM | GoPro PSNR/SSIM | HIDE PSNR/SSIM |
|---|---|---|---|---|---|
| WSA | 32.81/0.9021 | 29.01/0.7914 | 27.09/0.8138 | 32.09/0.9295 | 30.41/0.9129 |
| OCA(ours) | **32.82/0.9023** | **29.05/0.7925** | **27.17/0.8158** | **32.25/0.9316** | **30.50/0.9136** |

**Table 9:** Ablation study on different choices for TSAB and SSAB.

| Model | Params | Set5 PSNR/SSIM | Set14 PSNR/SSIM | Urban100 PSNR/SSIM | GoPro PSNR/SSIM | HIDE PSNR/SSIM |
|---|---|---|---|---|---|---|
| MDTA+Swin | 10.6M | <u>32.79</u>/**0.9018** | 29.00/0.7913 | 27.06/0.8130 | 31.67/<u>0.9204</u> | <u>30.32</u>/<u>0.9118</u> |
| MDTA+Dilation | 10.6M | <u>32.74</u>/0.9013 | 28.98/0.7904 | 27.00/0.8110 | 31.85/0.9266 | 30.22/0.9107 |
| CAB+OCA | 10.8M | 32.74/0.9016 | <u>29.04</u>/**0.7927** | **27.18/0.8165** | **32.47/0.9341** | <u>30.39</u>/0.9117 |
| MDTA+OCA(ours) | 10.6M | **32.82/0.9023** | **29.05**/<u>0.7925</u> | <u>27.17</u>/<u>0.8158</u> | <u>32.25</u>/**0.9316** | **30.50/0.9136** |

**Table 10:** Model Complexity Comparison. FLOPs are calculated for $256 \times 256$ input.

| Model | Params | FLOPs | Set5 PSNR/SSIM | Set14 PSNR/SSIM | Urban100 PSNR/SSIM | GoPro PSNR/SSIM | HIDE PSNR/SSIM |
|---|---|---|---|---|---|---|---|
| Restormer | 26.1M | 141.0G | 32.94/0.9039 | 29.06/0.7934 | 27.32/0.8199 | 32.92/0.9419 | 31.22/0.9226 |
| X-Restormer-s | 22.3M | 140.9G | <u>33.04</u>/**0.9047** | **29.18/0.7963** | <u>27.63</u>/**0.8287** | <u>33.41</u>/**0.9452** | <u>31.52</u>/**0.9277** |
| X-Restormer | 26.0M | 164.3G | **33.16/0.9058** | <u>29.17</u>/**0.7963** | **27.66/0.8291** | **33.44/0.9459** | **31.76/0.9299** |

**Table 11:** Comparisons of computational costs ($256 \times 256$).

| Method | MPRNet | SwinIR | Uformer | NAFNet | Restormer | X-Restormer |
|---|---|---|---|---|---|---|
| Params | 20.1M | 11.6M | 50.9M | 115.9M | 26.0M | 26.1M |
| FLOPs | 572.9G | 752.1G | 89.46G | 63.6G | 141.0G | 164.3G |
| runtime | 0.049s | 0.233s | 0.059s | 0.035s | 0.087s | 0.101s |

## 6  Model Complexity Comparison

In Tab. 10, we present the model complexity comparison of our X-Restormer with Restormer. For a clearer comparison, we also provide a variant X-Restormer-s, by reducing the original dimension of X-Restormer to 44. In Tab. 11, we provide comprehensive computational costs for different networks. For a fair comparison, we use the models trained on the **all-in-one** setting.

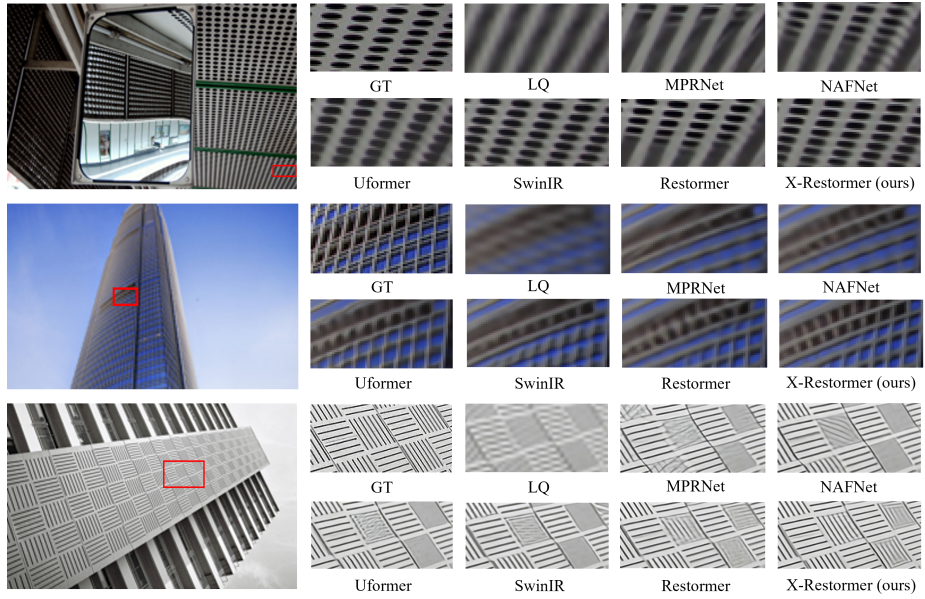

**Fig. 3:** Visual comparison on ×4 image SR.

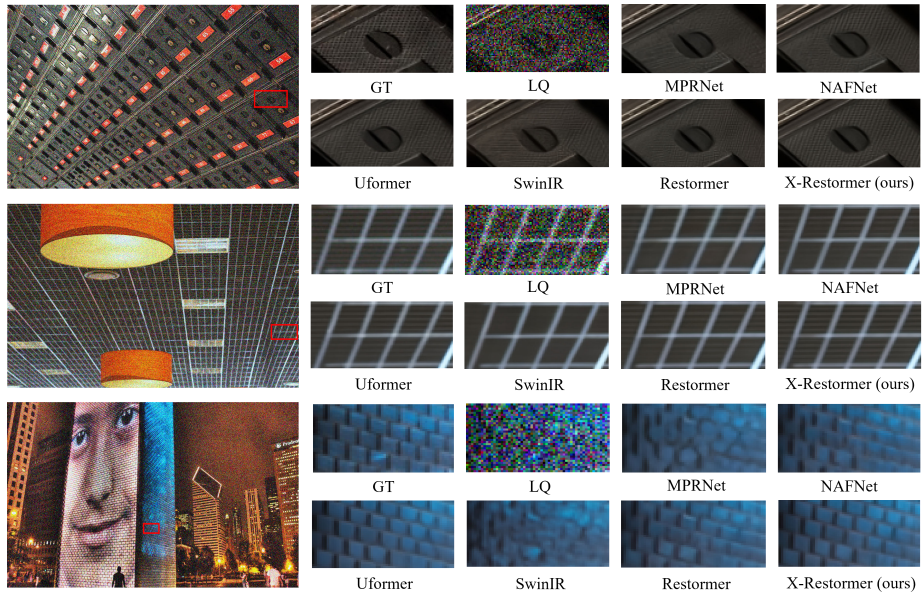

**Fig. 4:** Visual comparison on image denoising with the Gaussian noise level $\sigma = 50$.

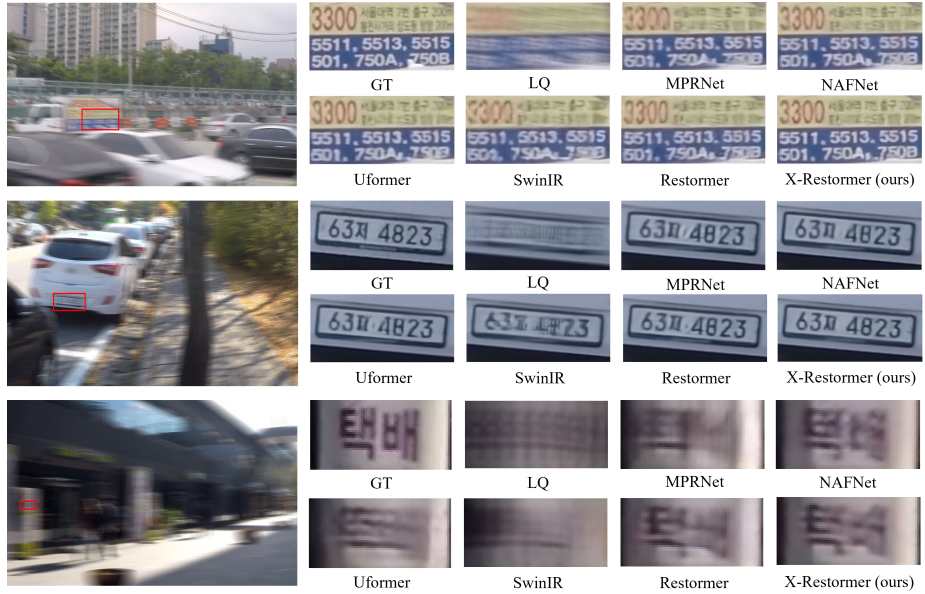

**Fig. 5:** Visual comparison on motion deblur.

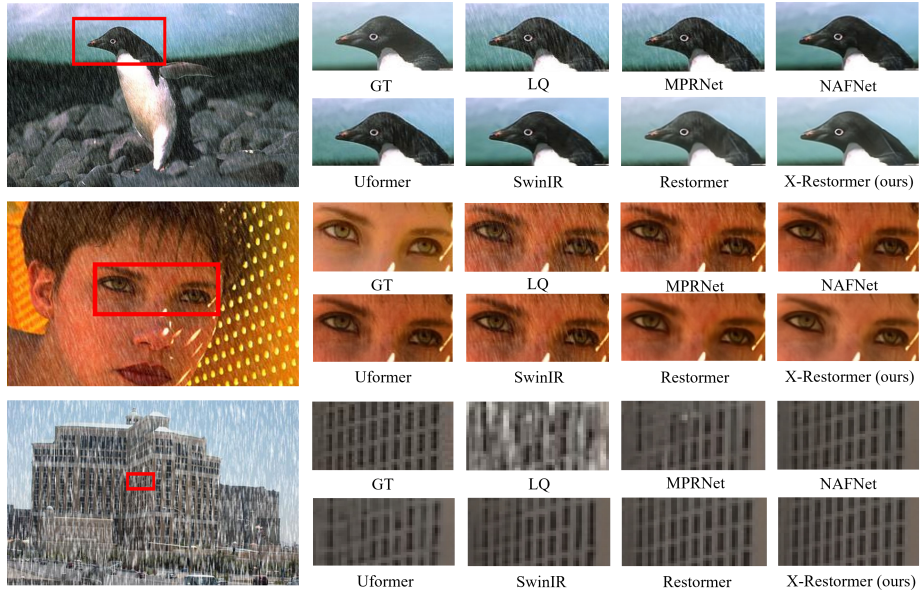

**Fig. 6:** Visual comparison on image deraining.

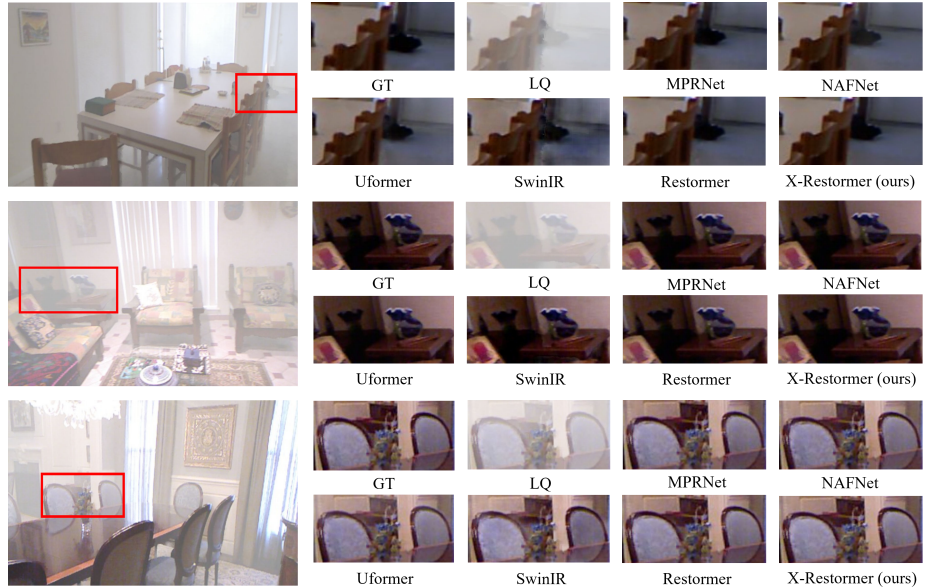

**Fig. 7:** Visual comparison on image dehazing.



\end{document}